\documentclass[runningheads]{llncs}

\usepackage{amsmath}
\usepackage{amsfonts}
\usepackage{epsfig}
\usepackage{subfigure}
\usepackage{amssymb}
\usepackage{graphicx}

\usepackage{url}
\urldef{\mailsa}\path|{bert,jebara}@cs.columbia.edu|

\begin{document}

\mainmatter  

\title{Approximating the Permanent with Belief Propagation}

\titlerunning{Approximating the Permanent with Belief Propagation}

%
%
\author{ Bert Huang
\and 
Tony Jebara
}

\newcommand{\trans}{^{\ensuremath{\mathsf{T}}}}
\def\todo{{\huge TO DO: }\textsf}
\authorrunning{Bert Huang, Tony Jebara}

\institute{
Computer Science Department\\
Columbia University\\
New York, NY 10027 \\
\mailsa\\
\url{http://www.cs.columbia.edu/learning}}

%
%

\toctitle{Approximating the Permanent with Belief Propagation}
\tocauthor{Bert Huang and Tony Jebara}

\maketitle

\begin{abstract}

This work describes a method of approximating matrix permanents efficiently using belief propagation. We formulate a probability distribution whose partition function is exactly the permanent, then use Bethe free energy to approximate this partition function. After deriving some speedups to standard belief propagation, the resulting algorithm requires $(n^2)$ time per iteration. Finally, we demonstrate the advantages of using this approximation.
\end{abstract}

\footnotetext{This work was done in late 2007 and early 2008.}

\section{Introduction}

The permanent is a scalar quantity computed from a matrix and has been
an active topic of research for well over a century. It plays a role
in cryptography and statistical physics where it is fundamental to
Ising and dimer models. While the determinant of an $n \times n$
matrix can be evaluated exactly in sub-cubic time, efficient methods
for computing the permanent have remained elusive. Since the permanent
is $\#$P-complete, efficient exact evaluations cannot be found in
general. The best exact methods improve over brute force (${\cal
O}(n!)$) and include Ryser's algorithm \cite{Ryser63,SerWan05} which
requires as many as $\Theta(n2^n)$ arithmetic operations. Recently,
promising fully-polynomial randomized approximate schemes (FPRAS) have
emerged which provide arbitrarily close approximations. Many of these
methods build on initial results by Broder \cite{Broder92} who applied
Markov chain Monte Carlo (a popular tool in machine learning and
statistics) for sampling perfect matchings to approximate the
permanent.  Recently, significant progress has produced an FPRAS that
can handle arbitrary $n \times n$ matrices with non-negative entries
\cite{JerSin03}. The method uses Markov chain Monte Carlo and only
requires a polynomial order of samples. 

However, while these methods have tight theoretical guarantees, they carry expensive constant factors, not to mention relatively high polynomial running times that discourage their usage in practical applications. In particular, we have experienced that the prominent algorithm in \cite{JerSin03} is slower than Ryser's exact algorithm for any feasible matrix size, and project that it only becomes faster around $n>30$. 

It remains to be seen if other approximate inference methods can be
brought to bear on the permanent.  For instance, loopy belief
propagation has also recently gained prominence in the machine
learning community. The method is exact for singly-connected networks
such as trees. In certain special loopy graph cases, including graphs
with a single loop, bipartite matching graphs \cite{BayShaSha05} and
bipartite multi-matching graphs \cite{HuaJeb07}, the convergence of BP
has been proven. In more general loopy graphs, loopy BP still
maintains some surprising empirical success.  Theoretical
understanding of the convergence of loopy BP has recently been
improved by noting certain general conditions for its fixed points and
relating them to minima of Bethe free energy. This article proposes
belief propagation for computing the permanent and investigates some
theoretical and experimental properties.

In Section \ref{sct:partition}, we describe a probability distribution
parameterized by a matrix similar to those described in \cite
{BayShaSha05,HuaJeb07} for which the partition function is exactly
the permanent. In Section \ref{sct:bethe}, we discuss Bethe free
energy and introduce belief propagation as a method of finding a
suitable set of pseudo-marginals for the Bethe approximation. In
Section \ref{sct:experiments}, we report results from experiments. We
then conclude with a brief discussion.

\section{The Permanent as a Partition Function}
\label{sct:partition}

Given an $n\times n$ non-negative matrix $W$, the matrix permanent is
\begin{equation}
\sum_{\pi \in S_{n}} \prod_{i=1}^n W_{i \pi(i)}.
\end{equation}
Here $S_{n}$ refers to the symmetric group on the set $\{1, \ldots,
n\}$, and can be thought of as the set of all permutations of the
columns of $W$. To gain some intuition toward the upcoming analysis,
we can think of the matrix $W$ as defining some function $f(\pi; W)$
over $S_{n}$.  In particular, the permanent can be rewritten as
\begin{eqnarray*}
\textrm{per}(W) = \sum_{\pi \in S_{n}} f(\pi; W), \hspace{.2in} \\
\textrm{ where }\hspace{.2in} 
f(\pi; W) = \prod_{i=1}^n W_{i \pi(i)}.\nonumber
\end{eqnarray*}
The output of $f$ is non-negative, so we consider $f$ a density 
function over the space of all permutations. 

If we think of a permutation as a
perfect matching or assignment between a set of $n$ objects $A$ and
another set of $n$ object $B$, we relax the domain by considering all
possible assignments of imperfect matchings for each item in the sets.

Consider the set of assignment variables $X = \{x_1, \ldots, x_n\}$,
and the set of assignment variables $Y = \{y_1, \ldots, y_n\}$, such
that $x_i, y_j \in \{1, \ldots, n\}, \forall i,j$. The value of the
variable $x_i$ is the assignment for the $i$'th object in $A$, in
other words the value of $x_i$ is the object in $B$ being selected
(and vice versa for the variables $y_j$).
\begin{eqnarray*} 
\phi(x_i) &=& \sqrt{W_{i x_i}}, \hspace{.2in}
\phi(y_j) = \sqrt{W_{y_j j}}, \hspace{.2in}\\
 \psi(x_i, y_j)&=&I(\neg(j = x_i \oplus i = y_j)).
\end{eqnarray*}
We square-root the matrix entries because the factor formula 
multiplies both potentials for the $x$ and $y$ variables. We
 use $I()$ to refer to an indicator function such that
$I(\textrm{true}) = 1$ and $I(\textrm{false}) = 0$. Then the $\psi$
function outputs zero whenever any pair $(x_i, y_j)$ have settings
that cannot come from a true permutation (a perfect
matching). Specifically, if the $i$'th object in $A$ is assigned to
the $j$'th object in $B$, the $j$'th object in $B$ must be assigned to
the $i$'th object in $A$ (and vice versa) or else the density function
goes to zero. Given these definitions, we can define the equivalent
density function that subsumes $f(\pi)$ as follows:
\begin{equation}
\hat{f}(X,Y) = \prod_{i,j} \psi(x_i,y_j) \prod_{k} \phi(x_k) \phi
(y_k).\nonumber
\end{equation}
This permits us to write the following equivalent formulation of the permanent:
$\textrm{per}(W) = \sum_{X,Y} f(X,Y)$.
Finally, if we convert density function $\hat{f}$ into a 
valid probability, simply add a normalization constant to it, producing:
\begin{equation}
\label{eqn:distribution}
p(X,Y) = \frac{1}{Z(W)}\prod_{i,j} \psi(x_i,y_j) \prod_{k} \phi(x_k) 
\phi(y_k).
\end{equation}
The normalizer or partition function $Z(W)$ is the sum of $f(X,Y)$ for
all possible inputs $X,Y$. Therefore, the partition function
of this distribution is the matrix permanent of $W$.

\section{Bethe Free Energy}
\label{sct:bethe}

To approximate the partition function, we use the Bethe free energy
 approximation. The Bethe free energy of our distribution given
a belief state $b$ is
\begin{eqnarray}
\label{eqn:bethe}
F_{Bethe} &=& -\sum_{ij} \sum_{x_i, y_j} b(x_i,y_j) \ln \psi(x_i,y_j)\phi(x_i)
\phi(y_j) \nonumber\\
&&+ \sum_{ij} \sum_{x_i, y_j} b(x_i,y_j) \ln b(x_i,y_j) 
\nonumber\\
&& -(n-1)\sum_{i}\sum_{x_i} b(x_i) \ln b(x_i) \nonumber\\
&&-(n-1)\sum_{j} \sum_
{y_j} b(y_j) \ln b(y_j)
\end{eqnarray}
The belief state $b$ is a set of pseudo-marginals that are only
locally consistent with each other, but need not necessarily achieve
global consistency and do not have to be true marginals of a single
global distribution.  Thus, unlike the distributions evaluated by the
exact Gibbs free energy, the Bethe free energy involves pseudo-marginals
that do not necessarily agree with a true joint distribution over the
whole state-space.  The only constraints pseudo-marginals of our
bipartite distribution obey (in addition to non-negativity) are the
linear local constraints:
\begin{eqnarray*}
\sum_{y_j} b(x_i,y_j) &=& b(x_i),  \hspace{.2in}
\sum_{x_i} b(x_i,y_j) = b(y_j),
\:\forall i,j, \nonumber\\
\sum_{x_i,y_j} b(x_i,y_j) &=& 1.
\end{eqnarray*}
The class of true marginals is a subset of the class of
pseudo-marginals. In particular, true marginals also obey the
constraint $ \sum_{X\setminus x} p(X) = p(x)$, which pseudo-marginals
are free to violate.

We will use the approximation
\begin{equation}
\textrm{per}(W) \approx \exp\left(- \min_{b} F_{\textrm{Bethe}}(b)\right)
\end{equation}

\subsection{Belief Propagation}

The canonical algorithm for (locally) minimizing the Bethe free energy
is {\em Belief Propagation}. We use the dampened belief propagation
described in \cite{Heskes2004a}, which the author derives as a (not
necessarily convex) minimization of Bethe free energy. Belief
Propagation is a message passing algorithm that iteratively updates
messages between variables that define the local beliefs. Let
$m_{x_i}(y_j)$ be the message from $x_i$ to $y_j$. Then the beliefs
are defined by the messages as follows:
\begin{equation}
b(x_i, y_j) \propto \psi(x_i,y_j) \phi(x_i) \phi(y_j) \prod_{k
\neq j} m_{y_k}(x_i) \prod_{\ell \neq i} m_{x_\ell}(y_j)\nonumber\\
\end{equation}
\begin{eqnarray}
\label{eqn:bpbeliefs}
b(x_i) &\propto&  \phi(x_i)\prod_{k} m_{y_k}(x_i),\:\:\:\:
b(y_j) \propto
\phi(y_j) \prod_{k} m_{x_k} (y_j)
\end{eqnarray}
In each iteration, the messages are updated according to the following
update formula:
\begin{equation}
\label{eqn:fullmessage}
m_{x_i}^{\textrm{new}}(y_j) = \sum_{x_i} \left[\phi(x_i) \psi
(x_i,y_j) \prod_{k\neq j} m_{y_k} (x_i)\right]
\end{equation}
Finally, we dampen the messages to encourage a smoother optimization in log-space.
\begin{equation}
\label{eqn:dampening}
\ln m_{x_i}(y_j) \leftarrow \ln m_{x_i}(y_j) + \epsilon \left[\ln m_{x_i}^
{\textrm{new}}(y_j) - \ln m_{x_i}(y_j)\right]
\end{equation}
We use $\epsilon$ as a dampening rate as in \cite{Heskes2004a} and
dampen in log space because the messages of BP are
exponentiated Lagrange multipliers of Bethe optimization \cite
{Heskes2004a,YedFreeWei05,Yuille2002}. We next derive faster updates
of the messages (\ref{eqn:fullmessage}) and the Bethe free energy
(\ref{eqn:bethe}) with some careful algebraic tricks.

\subsection{Algorithmic Speedups}

Computing sum-product belief propagation quickly for our distribution
is challenging since any one variable sends a message vector of length
$n$ to each of its $n$ neighbors, so there are $2n^3$ values to update
each iteration. One way to ease the computational load is to avoid
redundant computation. In Equation (\ref{eqn:fullmessage}), the only
factor affected by the value of $y_j$ is the $\psi$
function. Therefore, we can explicitly define the update rules based
on the $\psi$ function, which will allow us to take advantage of the
fact that the computation for each setting of $y_j$ is similar. When
$y_j\neq i$, we have
\begin{eqnarray}
m_{x_i y_j}^{\textrm{not}}& =& \left( \sum_{x_i
\neq j}\phi(x_i) \prod_{k\neq j} m_{y_k}(x_i)\right) \nonumber\\
&=&  \left( \sum_{x_i
\neq j}\phi(x_i) m_{y_{x_i} x_i}^{\textrm{match}} \prod_{k\neq j, k\neq x_i} m_{y_k x_i}^{\textrm{not}}\right).
\end{eqnarray}
When $y_j = i$,
\begin{eqnarray}
m_{x_i y_j}^{\textrm{match}} &=&  \left( \phi(x_i=j) 
\prod_{k\neq j} m_{y_k}(x_i=j)\right) \nonumber\\
&=& \left( \phi(x_i=j) \prod_{k\neq j} m_{y_k x_i}^{\textrm{not}}\right).
\end{eqnarray}
We have reduced the full message vectors to only two possible values: $m^{\textrm{not}}$ is the message for when the variables are not matched and $m^{\textrm{match}}$ is for when they are matched. We further simplify the messages by dividing both values by $m_{x_i y_j}^{\textrm{not}}$. This gives us
\begin{eqnarray}
m_{x_i y_j}^{\textrm{not}}&= & 1\nonumber\\
m_{x_i y_j}^{\textrm{match}} &=& \frac{\phi(x_i=j) \prod_{k\neq j} m_{y_k x_i}^{\textrm{not}}}{ \sum_{x_i
\neq j}\phi(x_i) m_{y_{x_i} x_i}^{\textrm{match}} \prod_{k\neq j, k\neq x_i} m_{y_k x_i}^{\textrm{not}}}\nonumber\\
&=& \frac{\phi(x_i=j) }{ \sum_{k
\neq j}\phi(x_i=k) m_{y_k x_i}^{\textrm{match}} } 
\end{eqnarray}
We can now define a fast message update rule that only needs to update one value between each variable.
\begin{equation}
m_{x_i y_j} \leftarrow \frac{1}{Z} \phi(x_i=j)/\sum_{k\neq j} \phi(x_i=k)m_{y_k x_i}
\end{equation}
We can rewrite the belief update formulas using these new messages.
\begin{eqnarray}
b(x_i=j, y_j=i)& =& \frac{1}{Z_{ij}} \phi(x_i) \phi(y_j)\nonumber\\
b(x_i \neq j, y_j\neq i)& =& \frac{1}{Z_{ij}} \phi(x_i) \phi(y_j) m_{y_{x_i} x_i} m_{x_{y_j} y_j}\nonumber\\ 
b(x_i) &= &\frac{1}{Z} \phi(x_i) m_{y_{x_i} x_i}, \nonumber\\
 b(y_j) &=& 
\frac{1}{Z} \phi(y_j) m_{x_{y_j} y_j} 
\end{eqnarray}
With the simplified message updates, each iteration takes ${\cal O} (n)$ operations per node, resulting in an algorithm that takes ${\cal O}(n^2)$ operations per iteration. We demonstrate experimentally that the algorithm converges to within a certain tolerance in a constant number of iterations with respect to $n$, so in practice the ${\cal O} (n^3)$ operations it takes to compute Bethe free energy is the asymptotic bottleneck of our algorithm.

\subsection{Convergence}

One important open question about this work is whether or not we can
guarantee convergence.  Empirically, by initializing belief propagation to various random points in the feasible space, we found BP still converged to the same solution. 
The max-product algorithm is guaranteed to
converge to the correct maximum matching \cite{BayShaSha05,HuaJeb07}
via arguments on the unwrapped computation tree of belief
propagation. The matching graphical model does not not meet the
sufficient conditions provided in \cite{Heskes2006} nor does our
distribution fit the criteria for non-convex convergence provided in
\cite{TatJor02} and \cite{Heskes2004}.

In our analysis, we have found that the Bethe free energy is certainly non-convex near the vertices of the distribution. That is, if we evaluate the Bethe free energy on pseudomarginals corresponding to exactly one matching, and take a tiny step in the direction of a non-adjacent matching vertex, Bethe free energy increases. On the other hand, when we initialize belief propagation such that the beliefs are at a vertex, BP moves away from the apparent local minimum and converges to the same solution as other initializations. This behavior implies that, while the Bethe free energy within the matching constraints is non-convex, it may still have a unique zero-gradient point despite not fitting the criteria in \cite{Heskes2004}, which exploit the strength of potentials. 

Since all our empirical evidence implies that BP
always converges, we suspect that we have not yet correctly analyzed
the true space traversed during optimization. In particular, the
distribution described by Equation \ref{eqn:distribution} is defined
over the set of all $n^n$ possible $X,Y$ states, while it is only
nonzero in $n!$ states. Any beliefs derived from belief propagation
obey similar constraints, so it is reasonable to suspect that careful
analysis of the optimization with special attention to the oddities of
the distribution could yield more promising theoretical
guarantees. 

However, without being rigorous, we can note that the matching constraints created by the $\psi$ functions enforce that the singleton beliefs are exactly the matched pairwise beliefs. 
This means we can think of these as entries in a doubly-stochastic matrix $B$.
\begin{eqnarray}
b(x_i=j,y_j=i) = b(x_i=j) = b(y_j = i) \equiv B_{ij}
\end{eqnarray}
Therefore it becomes clear that there is a strong connection to the Sinkhorn operation \cite{Sinkhorn67}, which iteratively scales rows and columns of a matrix until it converges to a doubly-stochastic matrix. It has been shown that the Sinkhorn operation effectively minimizes the pseudo-KL divergence between some matrix and the doubly-stochastic matrix\cite{Rangarajan}.
\begin{eqnarray*}
\min_{B}&& \sum_{ij} B_{ij}\log \frac{B_{ij}}{A_{ij}}\\
\textrm{s.t.}&& \sum_{i} B_{ij} = 1, \forall j, \:\:\: \sum_{j} B_{ij} = 1, \forall i
\end{eqnarray*}
Here the pseudo-KL divergence can be interpreted as the KL for each row and each column, each of which is an assignment distribution like in our matching setting. The Sinkhorn procedure is guaranteed to converge for indecomposable input matrices \cite{Sinkhorn67}, so the fact that the the procedure is reminiscent of ours is encouraging. However the two procedures differ enough that the guarantee does not directly translate. 

\begin{figure}[tb]
\begin{center}
\subfigure[Running time]{\epsfig{file=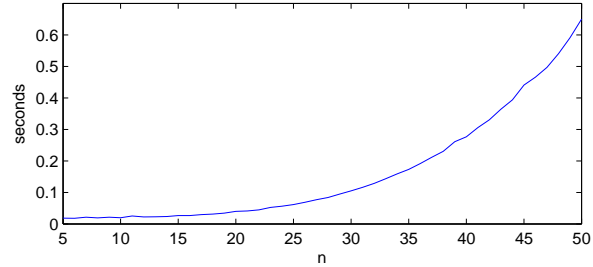, width=3.5in}}
\subfigure[Iterations]{\epsfig{file=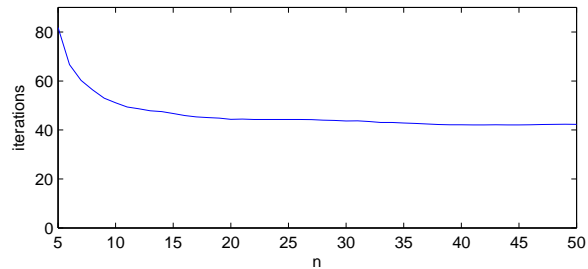, width=3.5in}}
\caption{\label{fig:runningtime} (a) Average running time until 
convergence of BP for $5\le n \le 50$. (b) Number 
of iterations.}
\end{center}
\end{figure}

\section{Experiments}
\label{sct:experiments}

In this section we evaluate the performance of this algorithm in 
terms of running time and accuracy, and finally we exemplify a 
possible usage of the approximate permanent as a kernel function.

\subsection{Running Time}

We ran belief propagation to approximate the permanents of random
matrices of sizes $n = [5, 50]$, recording the total running time and
the number of iterations to convergence. Surprisingly, the number of
iterations to convergence initially \emph{decreased} as $n$
grew, but appears to remain constant beyond $n>10$ or so. 
Running time still increased because the cost of updating each
iteration well subsumes the decrease in iterations to convergence.

In our implementation, we checked for convergence by computing the
absolute change in all the messages from the previous iteration, and
considered the algorithm converged if the sum of all the changes of
all $n^3$ messages was less than $1e-10$. In all cases, the resulting
beliefs were consistent with each other within comparable precision to
our convergence threshold. These experiments were run on a a 2.4 Ghz 
Intel Core 2 Duo Apple Macintosh running Mac OS X 10.5. The code is in 
$C$ and compiled using \texttt{gcc} version 4.0.1.

\begin{table}[htb]
\begin{center}
\caption{\label{tab:kendall} Normalized Kendall distances between 
the rankings of random matrices based on their true permanents and 
the rankings based on approximate permanents. See Figure \ref
{fig:accuracy} for plots of the approximations.}
\begin{tabular}{|c|c|c|c|c|} 
\hline
n & Bethe & Sampling & Det. & Diag. \\
\hline
10 & \textbf{0.00023} & 0.0248 & 0.3340 & 0.0724\\
\hline
8 & \textbf{0.0028} & 0.1285 & 0.4995 & 0.4057\\
\hline
5 & \textbf{0.0115} & 0.0914 & 0.4941 & 0.3834\\
\hline
\end{tabular}
\end{center}
\end{table}

\begin{figure}[tb]
\begin{center}
\epsfig{file=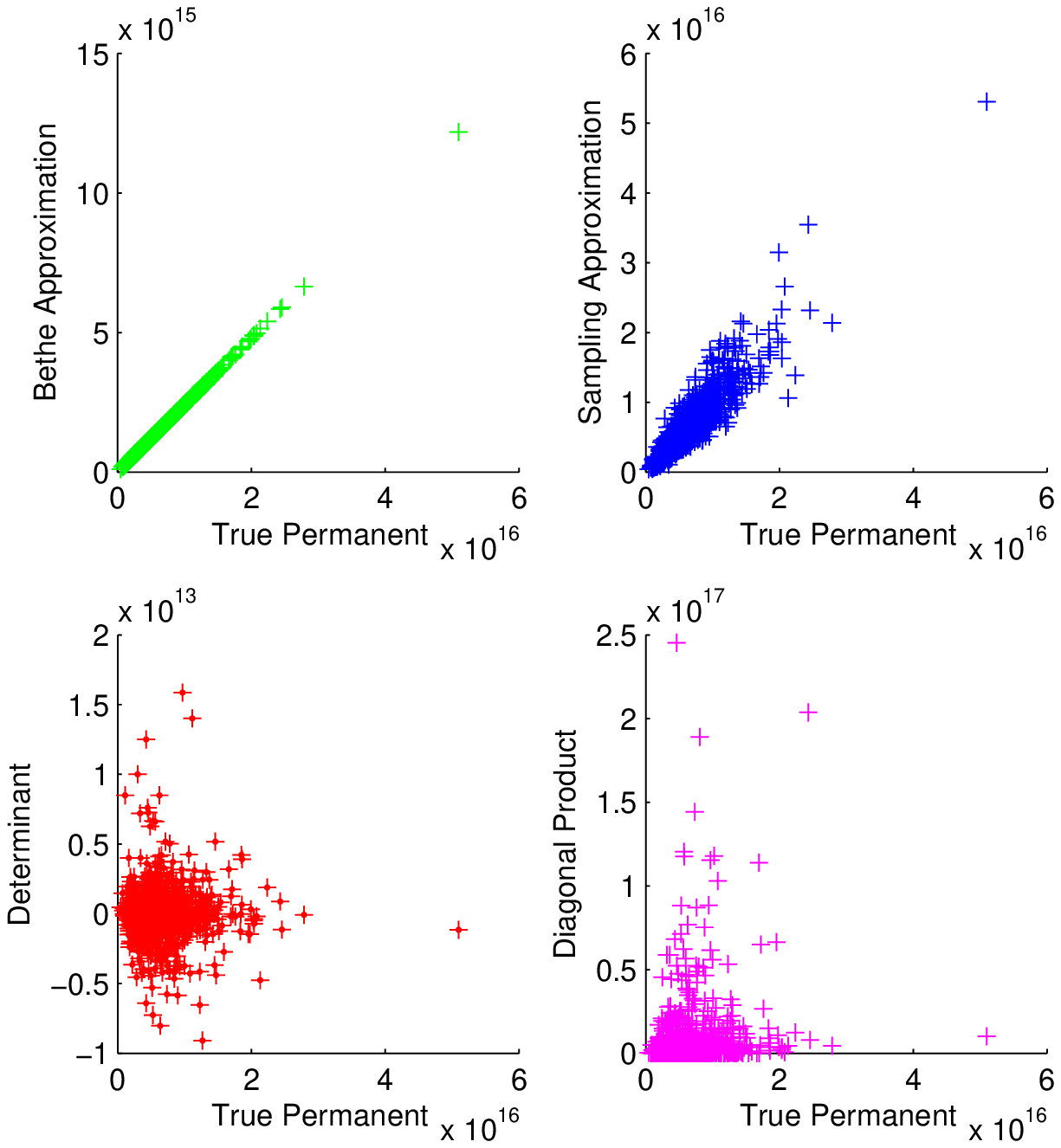, width=3.5in}
\caption{\label{fig:accuracy} Plots of the approximated permanent
versus the true permanent using four different methods. It is important to note that the scale of the y-axis varies from plot to plot. 
The diagonal
is extremely erratic and the determinant underestimates so much that
it is barely visible on the log scale. Sampling approximates values
much closer in absolute distance to the true permanent but does not
provide monotonicity in its approximations. Typically, this is more
important than absolute accuracy. Here we illustrate the results from
the $n=8$ case. We report results for $n=5$ and $10$ in Table
\ref{tab:kendall}.}
\end{center}
\end{figure}

\subsection{Accuracy of Approximation}

We evaluate the accuracy of our algorithm by creating 1000 random
matrices of sizes 5, 8 and 200 matrices of size 10. The entries of each 
of these matrices were randomly drawn from a uniform distribution in the interval $[0,50]$. 
We computed the
true permanents of these matrices, then computed approximate
permanents using our Bethe approximation. We also computed an
approximate using a naive sampling method, where we sample by choosing
random permutations and storing a cumulative sum of each permutation's
corresponding product. We sampled for the same amount of actual time
our belief propagation algorithm took to converge. Finally we also
computed two weak approximations: the determinant and the scaled
product of the diagonal entries.

In order to be able to compare to the true permanent, we had to limit
this analysis to small matrices. However, since MCMC sampling methods
such as in \cite{JerSin03} take ${\cal O}(n^{10})$ time to reach less
than some $\epsilon$ error, as matrix
size increases, the precision achievable in comparable time to our
algorithm would decrease. We scale the cumulative sum by
$\frac{n!}{s}$, where $s$ is the number of samples. This is the ratio
of the total possible permutations and the number of samples.

In our experiments, determinants and the products of diagonals are
neither accurate nor consistent approximations of the permanent. Sampling,
however, is accurate with respect to absolute distance to the
permanent, so for applications where that is most important, it may be
best to apply some sort of sampling method. Our Bethe approximation
seems the most consistent. While the approximations of the permanent are
off by a large amount, they seem to be consistently off by some
monotonic function of the true permanent. In many cases, this virtue
is more important than the absolute accuracy, since most applications
requiring a matrix permanent likely compare the permanents of
various matrices. These results are visualized for $n=8$ in Figure
\ref{fig:accuracy}.

To measure the monotonicity and consistency of these approximations,
we consider the Kendall distance \cite{fagin03comparing} between the
ranking of the random matrices according to the true permanent and
their rankings according to the approximations. Kendall distance is a
popular way of measuring the distance between two permutations. The
Kendall distance between two permutations $\pi_1$ and $\pi_2$ is
\begin{eqnarray*}
D_{\textrm{Kendall}}(\pi_1,\pi_2) &=& \sum_{i=1}^n \sum_{j=i+1}^n I\left((\pi_1(i)<\pi_1(j)) \wedge (\pi_2(i)> \pi_2(j))\right).
\end{eqnarray*}
In other words, it is the total number of pairs where $\pi_1$ and
$\pi_2$ disagree on the ordering. We normalize the Kendall distance by
dividing by $\frac{n(n-1)}{2}$, the maximum possible distance between
permutations, so the distance will always be in the range
$[0,1]$. Table \ref{tab:kendall} lists the Kendall distances between
the true permanent ranking and the four approximations. The Kendall
distance of the Bethe approximation is consistently less than that of
our sampler.

\begin{table}
\caption{\label{tab:svm} Left: Error rates of running SVM using
various kernels on the original three UCI datasets and data where the
features are shuffled randomly for each datum. Right: UCI resampled
pendigits data with order of points removed. Error rates of
1-versus-all multi-class SVM using various kernels.}
\subtable{\begin{tabular}{|l|c|c|c|}
\hline
\textbf{Kernel} & \textbf{Heart} & \textbf{Pima} & \textbf{Ion.}\\
\hline
\hline
Original Linear& 0.1600 & 0.2600 & 0.1640\\
\hline
Orig. RBF $\sigma=0.3$ & 0.2908 & 0.3160 & 0.1240\\
\hline
Orig. RBF $\sigma=0.5$ & 0.2158 & 0.3220 & 0.0760\\
\hline
Orig. RBF $\sigma=0.7$ & 0.1912 & 0.2760 & 0.0960\\
\hline
\hline
Shuffled Linear& 0.2456 & 0.3080 & 0.2640\\
\hline
Shuff. RBF $\sigma=0.3$ & 0.4742 & 0.3620 & 0.4840\\
\hline
Shuff. RBF $\sigma=0.5$ & 0.3294 & 0.3140 & 0.3580\\
\hline
Shuff. RBF $\sigma=0.7$ & 0.2964 & 0.3280 & 0.2700\\
\hline
\hline
Bethe $\sigma=0.3$ & 0.2192 & \textbf{0.2900} & \textbf{0.1000}\\
\hline
Bethe $\sigma=0.5$ & \textbf{0.2140} & \textbf{0.2900} & 0.1380\\
\hline
Bethe $\sigma=0.7$ & 0.2164 & 0.2920 & 0.1380\\
\hline
\end{tabular}}
\hspace{.15in}
\subtable{\begin{tabular}{|l|c|}
\hline
\textbf{Kernel} & \textbf{PenDigits}\\
\hline
\hline
Sorted Linear& 0.3960\\
\hline
Sorted RBF $\sigma=0.2$ & 0.4223\\
\hline
Sorted RBF $\sigma=0.3$ & 0.3407\\
\hline
Sorted RBF $\sigma=0.5$ & 0.3277\\
\hline
\hline
Shuffled Linear& 0.7987\\
\hline
Shuff. RBF $\sigma=0.2$ & 0.9183\\
\hline
Shuff. RBF $\sigma=0.3$ & 0.9120\\
\hline
Shuff. RBF $\sigma=0.5$ & 0.8657\\
\hline
\hline
Bethe $\sigma=0.2$ & 0.1463\\
\hline
Bethe $\sigma=0.3$ & \textbf{0.1190}\\
\hline
Bethe $\sigma=0.5$ & 0.1707\\
\hline
\end{tabular}}
\end{table}

\subsection{Approximate Permanent Kernel}

To illustrate a possible usage of an efficient permanent
approximation, we use a recent result \cite{Cut07} proving that the
permanent of a valid kernel matrix between two sets of points is also
a valid kernel between point sets. Since the permanent is invariant to
permutation, we decided to try a few classification tasks using an
approximate permanent kernel. The permanent kernel is computed by
first computing a valid subkernel between a pairs of elements in two
sets, then the permanent of those subkernel evaluations is taken as
the kernel value between the data. Surprisingly, in experiments the kernel matrix
produced by our algorithm was a valid positive definite matrix. This
discovery opens up some intriguing questions to be explored later.

We ran a similar experiment to \cite{ShiJeb06} where we took a the
first 200 examples of each of the Cleveland Heart Disease, Pima
Diabetes, and Ionosphere datasets from the UCI repository \cite{UCI},
and randomly permuted the features of each example, then compare the
result of training an SVM on this shuffled data. We also provide the
performance of the kernels on the unshuffled data for
comparison. After normalizing the features of the data to the
$[0,1]^D$ box, we chose three reasonable settings of $\sigma$ for the
RBF kernels and cross validated over various settings of the
regularization parameter $C$. We used RBF kernels between pairs of
data as the permanent subkernel. Finally, we report the average error
over 50 random splits of 150 training points and 50 testing
points. Not surprisingly, the permanent kernel is robust to the
shuffling and outperforms the standard kernels.

We also tested the Bethe kernel on the pendigits dataset, also from
the UCI repository. The original pendigits data consists of
stylus coordinates of test subjects writing digits. We used the
preprocessed version that has been resampled spatially and
temporally. However, we omit the order information and treat the input
as a cloud of unordered points. Since there is a natural spatial
interpretation of this data, so we compare to sorting by distance from
origin, a simple method of handling unordered data. We chose slightly
different $\sigma$ values for the RBF kernels. For this dataset, there
are 10 classes, one for each digit, so we used a one-versus-all
strategy for multi-class classification. Here we averaged over only 10
random splits of 300 training points and 300 testing points (see Table \ref{tab:svm}).

Based on our experiments, the permanent kernel typically does not
outperform standard kernels when permutation invariance is not
inherently necessary in the data, but when we induce the necessity of
such invariance, its utility becomes clear.

\section{Discussion and Future Directions}

We have described an algorithm based on BP over
a specific distribution that allows an efficient approximation of the
$\#P$ matrix permanent operation. We write a probability distribution over matchings and use Bethe free energy to approximate the partition function of this distribution. 
The algorithm is significantly faster than sampling methods, but attempts to minimize a function that approximates the permanent. Therefore it is limited by the quality of the Bethe approximation so it cannot be run longer to obtain a better approximation like sampling methods can. However, we have shown that even on small matrices where sampling methods can achieve extremely high accuracy of approximation, our method is more well behaved than sampling, which can approach the exact value from above or below.

In the future, we can try other methods of approximating the partition function such as generalized belief propagation \cite{YedFreeWei05},
which takes advantage of higher order Kikuchi approximations of free
energy. Unfortunately the structure of our graphical model causes
higher order interactions to become expensive quickly, since each
variable has exactly $N$ neighbors. Similarly, the bounds on the
partition function in \cite{WaiJaaWil02} are based on
spanning subtrees in the graph, and again the fully connected
bipartite structure makes it difficult to capture the true behavior of
the distribution with trees.

Finally, the positive definiteness of the kernels we
computed is surprising, and requires further analysis. The exact permanent of a valid kernel
forms a valid Mercer kernel \cite{Cut07} because it is a sum of positive products, but 
since our algorithm outputs the results of an iterative approximation of the permanent, it 
is not obvious why the resulting output would obey the positive definite constraints.

\subsubsection*{Acknowledgments}

{\small
\bibliographystyle{plain}
\bibliography{permanentTR}
}

\end{document}